\documentclass[10pt,twocolumn,letterpaper]{article}

\usepackage{cvpr}              

\usepackage{graphicx}
\usepackage{amsmath}
\usepackage{amssymb}
\usepackage{booktabs}
\usepackage{rotating}
\usepackage{multirow}

\usepackage[pagebackref,breaklinks,colorlinks]{hyperref}

\usepackage[capitalize]{cleveref}
\crefname{section}{Sec.}{Secs.}
\Crefname{section}{Section}{Sections}
\Crefname{table}{Table}{Tables}
\crefname{table}{Tab.}{Tabs.}

\def\cvprPaperID{6499} 
\def\confName{CVPR}
\def\confYear{2023}

\begin{document}

\title{Texture-guided Saliency Distilling for Unsupervised Salient Object Detection}

\author{Huajun~Zhou$^{1}$,
        Bo~Qiao$^{1}$,
        Lingxiao~Yang$^{1}$,
        Jianhuang~Lai$^{1,2,3}$,
        Xiaohua~Xie$^{1,2,3}$\thanks{Corresponding author. This project is supported by the Key-Area Research and Development Program of Guangdong Province (2019B010155003), the National Natural Science Foundation of China (U22A2095, 62072482, 62076258, 62206316), and the Guangdong NSF Project (2022A1515011254). }
        \\
        $^{1}$School of Computer Science and Engineering, Sun Yat-sen University, China\\
        $^{2}$Guangdong Province Key Laboratory of Information Security Technology, China\\
        $^{3}$Key Laboratory of Machine Intelligence and Advanced Computing, Ministry of Education, China\\
        }

\maketitle

\begin{abstract}
Deep Learning-based Unsupervised Salient Object Detection (USOD) mainly relies on the noisy saliency pseudo labels that have been generated from traditional handcraft methods or pre-trained networks. To cope with the noisy labels problem, a class of methods focus on only easy samples with reliable labels but ignore valuable knowledge in hard samples. In this paper, we propose a novel USOD method to mine rich and accurate saliency knowledge from both easy and hard samples. First, we propose a Confidence-aware Saliency Distilling (CSD) strategy that scores samples conditioned on samples’ confidences, which guides the model to distill saliency knowledge from easy samples to hard samples progressively. Second, we propose a Boundary-aware Texture Matching (BTM) strategy to refine the boundaries of noisy labels by matching the textures around the predicted boundaries. Extensive experiments on RGB, RGB-D, RGB-T, and video SOD benchmarks prove that our method achieves state-of-the-art USOD performance. Code is available at \url{www.github.com/moothes/A2S-v2}.
\end{abstract}

\section{Introduction}
\label{sec:intro}
Unsupervised Salient Object Detection (USOD) methods aim to correctly localize and precisely segment salient objects simultaneously without using manual annotations.
Compared to the supervised methods, USOD methods can easily adapt to more practical scenarios (\textit{e.g.}, industrial or medical images) where a large number of labeled images may be very hard to collect.
Moreover, USOD methods also can assist some related methods for other tasks, \textit{e.g.}, object recognition \cite{rec1,rec2} and object detection \cite{det1,det2}. 
However, diverse objects, complex backgrounds, and other challenging conditions bring severe challenges to USOD methods.

\begin{figure}[!t]
\centering
\begin{minipage}{0.49 \textwidth}
\subfloat[Image]{
\includegraphics[width=0.78in,height=0.6in]{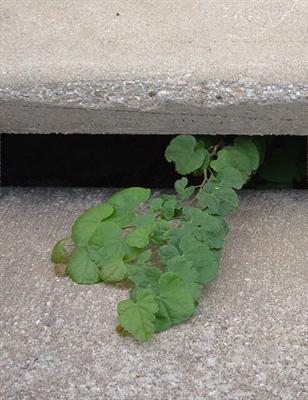}}
\subfloat[GT]{
\includegraphics[width=0.78in,height=0.6in]{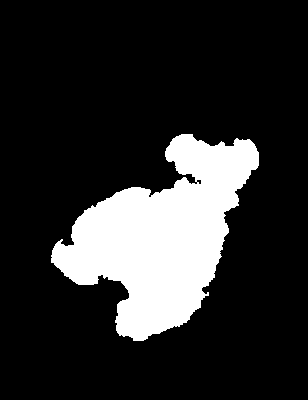}}
\subfloat[RBD]{
\includegraphics[width=0.78in,height=0.6in]{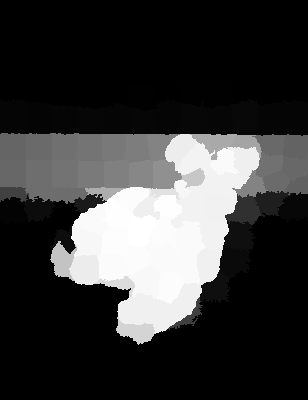}}
\subfloat[HS]{
\includegraphics[width=0.78in,height=0.6in]{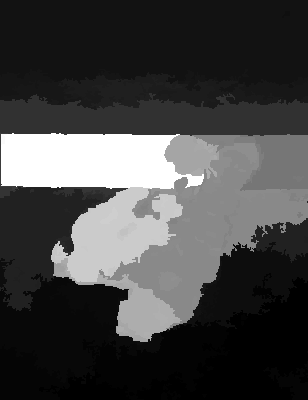}}
\end{minipage}

\begin{minipage}{1 \textwidth}
\subfloat[USPS]{
\includegraphics[width=0.78in,height=0.6in]{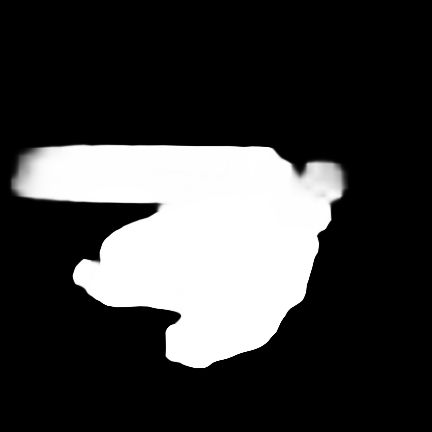}}
\subfloat[A2S]{
\includegraphics[width=0.78in,height=0.6in]{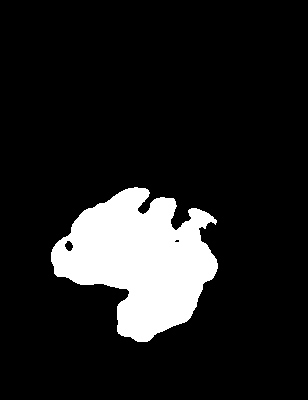}}
\subfloat[Activation]{
\includegraphics[width=0.78in,height=0.6in]{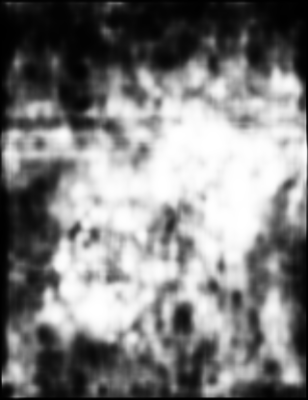}}
\subfloat[Ours]{
\includegraphics[width=0.78in,height=0.6in]{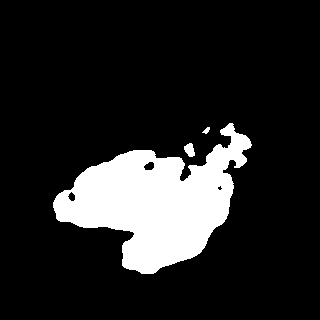}}
\end{minipage}

\caption{Saliency maps generated by USOD methods. Our result is generated based on the activation map of a deep network.}
\label{fig:pseudo}
\vspace{-0.15in}
\end{figure}

Most Deep Learning-based (DL-based) methods \cite{sbf,mnl,usps,dcfd,dsu,msum} base on the saliency cues extracted by traditional SOD methods (Fig. \ref{fig:pseudo}-c and \ref{fig:pseudo}-d).
These handcrafted features related cues are employed as pseudo labels to train deep networks under certain constraints, \textit{e.g.}, binary cross-entropy (BCE) loss.
However, saliency cues by traditional methods usually shift away from target objects, especially in complex scenes.
Moreover, conventional constraints, such as BCE loss, works well on fully-supervised SOD methods, but is suboptimal when fitting the noisy labels for unsupervised methods (Fig. \ref{fig:pseudo}-e).
Recently, Zhou et al. \cite{a2s} addressed the first issue by extracting saliency cues (Fig. \ref{fig:pseudo}-f) based on a unsupervisedly pre-trained network (Fig. \ref{fig:pseudo}-g) instead of using traditional methods.
During training, they focus on learning reliable saliency knowledge from easy samples, but ignore latent knowledge in hard samples.
The main reason is that hard samples may be wrongly-labeled and corrupt the fragile saliency knowledge learned in the early training phase. 
Therefore, to leverage hard samples, we argue that all samples should be employed in a meaningful order (i.e., from high reliable to low reliable), which is crucial for mining accurate knowledge from noisy labels.
Trained by such a strategy, the network can mine valuable knowledge from hard examples without corrupting the knowledge learned from easy samples.

Deep networks can learn to localize salient regions from noisy labels \cite{a2s}, but still struggle to find the precise boundaries of target objects.
Generally, the appearance around saliency boundary has a similar texture as in saliency map.
Therefore, matching the textures between different maps can serve as a guidance for producing reasonable saliency boundaries.
We will demonstrate that above strategies are applicable to multimodal data besides RGB image, including depth map, thermal image, and optical flow.

Based on the above analysis, we propose a novel framework to tackle the Unsupervised Salient Object Detection (USOD) tasks.
Specifically, two strategies are proposed to mine saliency knowledge from noisy saliency labels.	
First, we propose a Confidence-aware Saliency Distilling (CSD) scheme that scores samples with noisy labels conditioned on samples' confidences.
Then, our CSD guides the network to learn saliency knowledge from easy samples to more complex ones progressively by employing an adaptive loss conditioned on the training progress.
Second, we propose a Boundary-aware Texture Matching (BTM) strategy to refine the saliency boundaries of noisy labels by matching the textures around the predicted boundaries.
During training, the predicted saliency boundaries are shifting toward surrounding edges in the appearance space of the whole image.
Finally, guided by above two mechanisms, our method can produce high-quality pseudo labels to train generalized saliency detectors. 
Extensive experiments on RGB, RGB-D, RGB-T, and video SOD benchmarks prove that our method achieves state-of-the-art performance compared to existing USOD methods.

The main contributions of our novel USOD method are:
\begin{enumerate}
\item We propose a Confidence-aware Saliency Distilling (CSD) to mine rich and accurate saliency knowledge from noisy labels, which breaks through the limitation that existing methods cannot utilize hard samples.
\item We propose a Boundary-aware Texture Matching (BTM) to refine the boundary of the predicted saliency maps by matching textures in different spaces.
\item Extensive experiments on RGB, RGB-D, RGB-T and video SOD benchmarks prove that our method achieves state-of-the-art USOD performance.
\end{enumerate}

\section{Related Works}
\subsection{Supervised Salient Object Detection}
Researchers have developed a large family of fully-supervised Salient Object Detection (SOD) algorithms \cite{picanet,cpd,afnet,cag,page,pagrn,ucf,dss,amulet,basnet,gate,itsd,contour} in the past decades.
Ronneberger et al. \cite{unet} proposed a U-shape structure that progressively upsamples and concatenates the smaller features to the larger ones.
To ease the annotation burden, 
Zhang et al. \cite{wssa} relabeled the DUTS-TR dataset \cite{duts} with scribbles and leveraged an edge detection model for boundary localization.
Yu et al. \cite{lsc} proposed a local coherence loss to find precise boundary based on scribbles annotations.

Multimodal SOD tasks aim at using other modality data to improve the SOD performance, such as depth map, thermal image and optical flow.
Recently, abundant methods \cite{dsn2,ssav,adf,dsa2f,spnet,digr,pcsa,tenet,stvs,mied,midd} were proposed.
Most of these methods employ a two-stream encoder-decoder structure to aggregate multi-level information in multimodal data.
To reduce the annotating cost, Zhao et al. \cite{wvsod} proposed a video SOD dataset with scribbles annotations to indicate the location of salient objects.

Although the above methods have achieved significant performance, they require numerous human annotations for training, which are expensive to collect.

\begin{figure*}[!t]
\centering
\includegraphics[width=1 \textwidth]{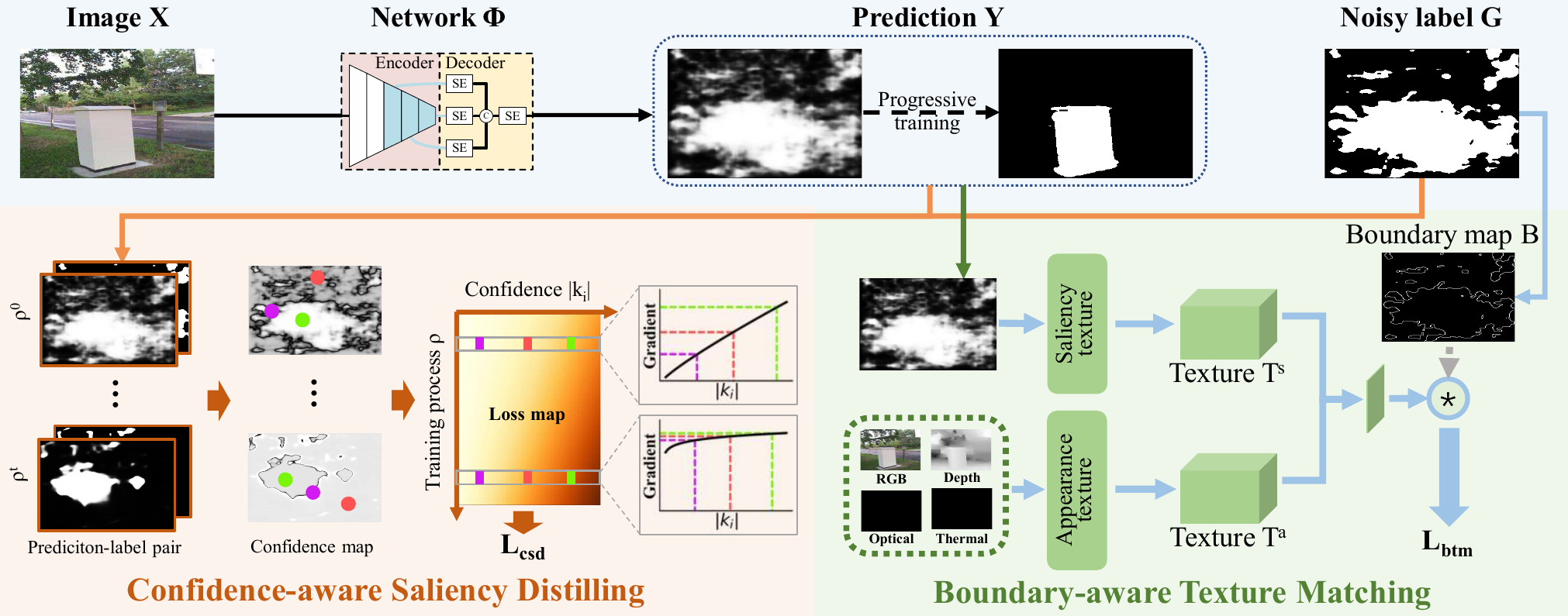}\\
\vspace{-0.1in}
\caption{Overview of the proposed unsupervised framework. 
The network are guided by two mechanisms: (1) Condidence-aware Saliency Distilling mines rich and accurate saliency knowledge from noisy labels, and (2) Boundary-aware Texture Matching makes the boundary of the predicted saliency map more accurate. We pad some constant maps for the missing modalities.}
\vspace{-0.1in}
\label{fig:framework}
\end{figure*}

\subsection{Unsupervised Salient Object Detection}
Traditional SOD methods \cite{dsr, mc, rbd, cssd} extracted saliency cues from images by modeling the correlation of hand-crafted features.
Inspired by the centralization prior, Jiang et al. \cite{mc} considered the distances between boundary superpixels and non-boundary superpixels as saliency scores. 
Yan et al. \cite{cssd} employed a tree-structure graphical model to compute the saliency results based on several over-segmented maps.
The above methods fail to accurately localize salient objects because the global information in hand-crafted features are not representative enough.

Existing DL-based USOD methods can be divided into two pipelines based on the method used to extract saliency cues from images.
First, most USOD methods \cite{sbf,mnl,usps,edns,dcfd,msum} focused on refining the coarse saliency cues extracted by several traditional SOD methods.
For example, 
Zhang et al. \cite{sbf} weighted these saliency cues by combining both intra-image and inter-image fusion streams.
Zhang et al. \cite{mnl} designed a noise modeling module to deal with noises in these saliency cues.
Nguyen et al. \cite{usps} used these saliency cues as labels to train multiple deep networks.
Ji et al. \cite{dsu} refined the saliency maps extracted by traditional SOD methods to produce more accurate saliency predictions.
Second, to prevent the localization errors caused by traditional SOD methods, Zhou et al. \cite{a2s} proposed a novel framework that converts the activation maps of a pre-trained network to high-quality pseudo labels.

In this paper, we propose two novel mechanisms to tackle USOD tasks and explore the potential of improving the quality of pseudo labels by using multimodal data.

\section{Our Approach}
In our method, we propose two novel strategies to mine accurate saliency knowledge based on the noisy activation maps generated by a deep network $\Phi$, as shown in Fig. \ref{fig:framework}.

\textbf{Activation map generation.}
Following previous work \cite{a2s}, we generate an activation map $Y$ for input image $X$ using a deep network $\Phi(X) = Y$:
\begin{equation}
\label{eqn:phi}
\begin{split}
E_3, E_4, E_5 &= Encoder(X),\\
F_i &= SE(E_i), i \in \{3,4,5\},\\
H &= SE(concat(F_3, F_4, F_5)),\\
Y &= inv(Sigmoid(sum(H - \bar H))),
\end{split}
\end{equation}
where $\bar H$ is the spatial mean of $H$.
Specifically, we employ the ResNet-50 \cite{resnet} pre-trained by MoCo-v2 \cite{mocov2} as our encoder, which is trained without extra manual annotations.
In our decoder, four SE blocks \cite{senet} integrate multiple encoder features into $H$.
After that, we set $H - \bar H$ to ensure that (1) noisy labels are adaptive to input images when using a fixed threshold and (2) coexisting of positive and negative samples.
Next, features are summed over the channel dimension to produce one-channel activation map.
Finally, the $Sigmoid$ function produces the final saliency scores, and the $inv$ function identifies the regions with more corner pixels as background.

\begin{figure*}[t]
\centering
\includegraphics[width=0.95 \textwidth]{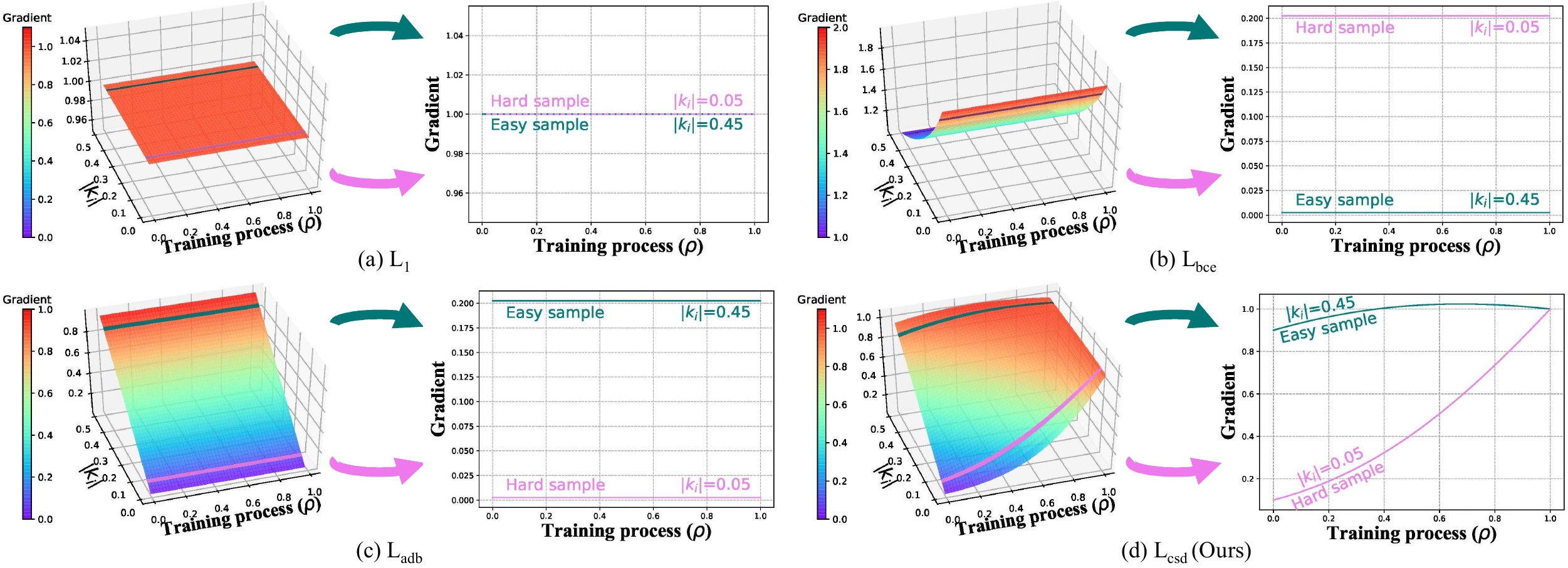}
\vspace{-0.1in}
\caption{The gradient landscapes of different loss functions.}
\vspace{-0.1in}
\label{fig:doloss}
\end{figure*}

\textbf{Training strategy.}
The activation maps produced by network $\Phi$ perceive some discriminative regions in input images \cite{a2s}, but still are low-quality because of widespread noises.
To improve the quality, we employ three strategies to train the network $\Phi$ by:
\begin{equation}
\label{eqn:lambda}
L_{\Phi} = \lambda_c (L_{csd} + \hat L_{csd}) + \lambda_b (L_{btm} + \hat L_{btm}) +\lambda_m L_{ms},
\end{equation}
where all $\lambda$ are hyperparameters.
$\hat{L}$ is the loss for the predictions of a different scale.
First, we propose a Confidence-aware Saliency Distilling (CSD) scheme to excavate valuable saliency knowledge from simple examples to more complex ones.
Second, we propose a Boundary-aware Texture Matching (BTM) strategy to align the boundaries in appearance and the predicted saliency maps.
In addition, a multi-scale consistency loss $L_{ms}$ ensures our method to produce consistent predictions for multi-scale inputs.

\subsection{Confidence-aware Saliency Distilling}
The primary challenge for USOD methods is how to localize salient regions.
Existing DL-based USOD methods \cite{sbf,usps,a2s} are based on noisy saliency cues from traditional SOD methods \cite{hs,mc,dsr,rbd} or pre-trained networks \cite{mocov2,resnet}.
Following \cite{a2s}, we employ the activation maps produced by a pre-trained network as our saliency cues instead of the saliency predictions of traditional SOD methods.
The main reasons are two-fold: (1) traditional methods require extra computation loads; (2) network can focus on mining saliency knowledge instead of fitting the inductive bias of traditional methods.
We binarize the saliency cues as the initial labels using a fixed threshold of 0.5, while Eqn. \ref{eqn:phi} ensures that these labels remain adaptive to the input image.

To generate high-quality pseudo labels, USPS \cite{usps} uses these initial labels to train deep networks like other fully-supervised methods.
However, the simple training strategy in fully-supervised methods is suboptimal for unsupervised methods.
Specifically, we define pixels with saliency scores close to 0.5 as hard examples because of their low confidences.
In the noisy labels, hard examples are likely to be wrongly labeled.
Therefore, it is difficult to learn robust saliency knowledge from these hard samples using traditional loss functions (\textit{e.g.}, BCE loss).
To this end, A2S \cite{a2s} learns reliable saliency knowledge from easy samples.
However, for hard examples, the saliency knowledge hidden in noises is not fully explored.
In summary, a customized strategy that organizes the samples in a more meaningful order is crucial for USOD methods.

To address the above problem, we propose a Confidence-aware Saliency Distillation (CSD) scheme that scores samples with noisy labels conditional on their confidences and training progress.
Concretely, easy samples contain reliable knowledge, and thus can assist our method to learn reliable saliency knowledge.
On the contrary, the saliency knowledge in hard samples is hidden within noises, and may corrupt the fragile saliency patterns in the early stages of network training.
Inspired by self-paced learning \cite{SPL}, we dynamically adjust the gradients for samples by introducing a factor $\rho$, which is linearly increasing from 0 to 1 as training proceeding.
Our $L_{csd}$ loss can be formulated as:
\begin{equation}
\label{eqn:csd}
L_{csd} = -\frac{1}{N} \sum_{i}^N |\Phi (\textbf{x}_i) - 0.5|^{(2^{1-\rho})},
\end{equation}
where $N$ is the number of pixels, and $\Phi (\textbf{x}_i)$ is the saliency prediction of pixel $\textbf{x}_i$.
Notice that noisy label G is omitted in Eqn. \ref{eqn:csd} because it is generated based on $\Phi (\textbf{x}_i)$.
Thus, we can calculate the confidence score $|k_i|$ of pixel $\textbf{x}_i$ in a simpler way as: $|k_i| = |\Phi (\textbf{x}_i) - 0.5|$, where the constant value 0.5 is related to $\bar H$ in Eqn. \ref{eqn:phi}.
The partial derivative of our $L_{csd}$ over $\Phi (\textbf{x}_i)$ is:
\begin{equation}
        \frac{\partial L_{csd}}{\partial \Phi (\textbf{x}_i)}  = -sign(k_i)2^{(1-\rho)}|k_i|^{2^{(1-\rho)}-1},
\end{equation}
where $sign(k_i) \in \{-1, 1\}$ for negative and positive values, respectively.

For an intuitive comparison, we draw the gradient landscapes of different losses in Fig. \ref{fig:doloss}, including $L_{1}$, $L_{bce}$, $L_{adb}$ \cite{a2s} and our $L_{csd}$.
The gradient of other losses are consistent throughout the training process, while our $L_{csd}$ varies with the training process.
Specifically, in the beginning, our $L_{csd}$ assigns low gradients to hard samples to learn reliable saliency knowledge from easy samples.
As training proceeding, the gradients of hard samples are increasing to mine more valuable saliency knowledge.

\begin{figure}[!t]
\centering
\includegraphics[width=0.48 \textwidth, height=1.5in]{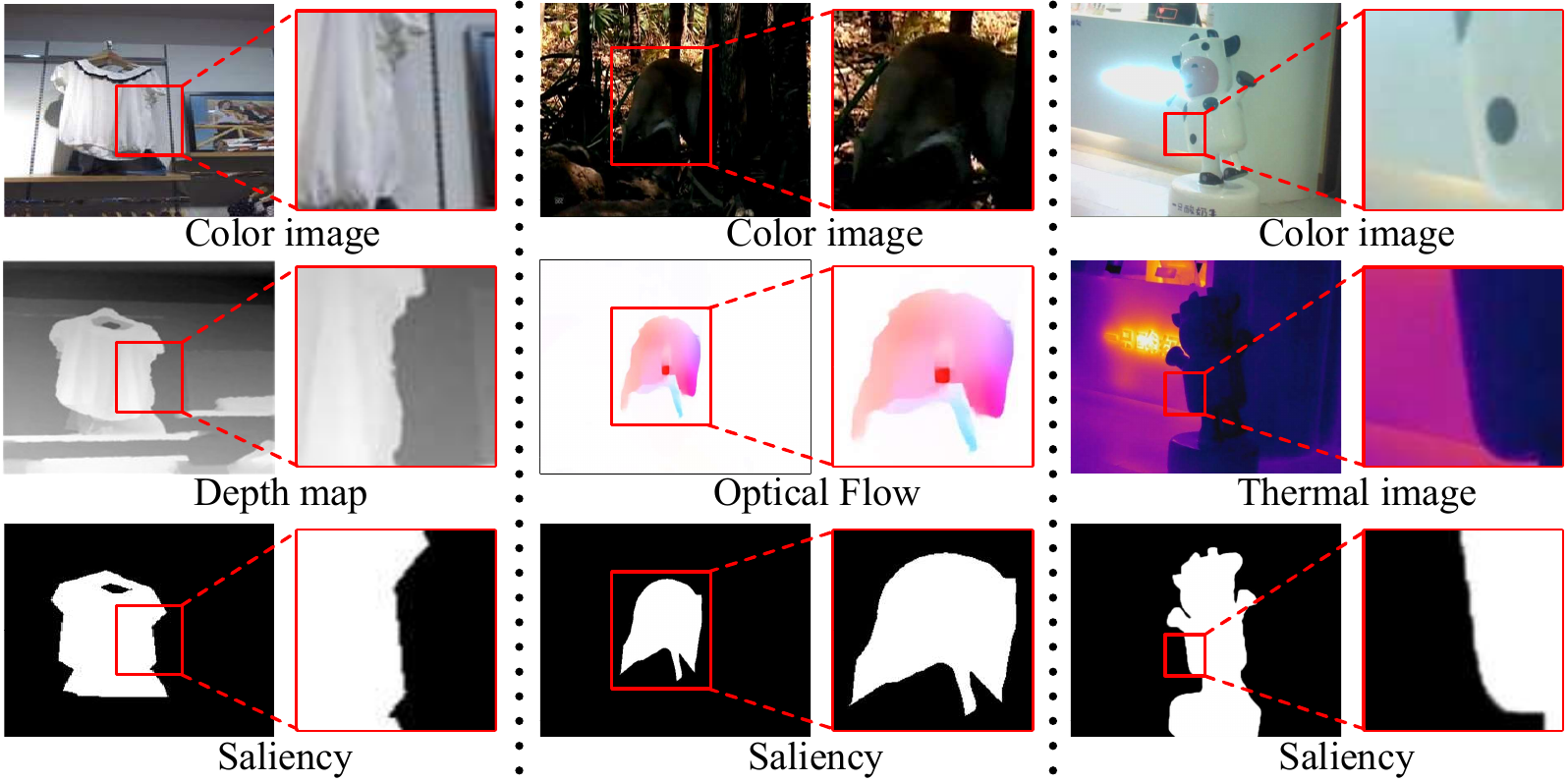}
\vspace{-0.25in}
\caption{Appearance information in multimodal data.}
\vspace{-0.1in}
\label{fig:multimodal}
\end{figure}

\subsection{Boundary-aware Texture Matching}
In general, the appearance of saliency boundary has a similar texture as in saliency prediction.
Therefore, matching the textures between different maps can guide our method to produce reasonable saliency scores.
This strategy is also applicable to other modalities, such as depth map, thermal image, and optical flow, as shown in Fig. \ref{fig:multimodal}.
By aggregating the rich appearance information in multimodal data, we can provide a more generalized guidance for our USOD method.

Based on the above analysis, we propose a Boundary-aware Texture Matching (BTM) strategy to align the saliency boundaries and image edges by matching their textures.
First of all, similar to the LBP feature \cite{lbp}, we extract texture vectors from saliency predictions and appearance, respectively.
For saliency predictions, the texture vector $T^s_i = [t^s_{i, 1}, t^s_{i, 2}, ..., t^s_{i, k^2}]$ for the $i$-th pixel is computed by $t^s_{i, j} = |\Phi (\textbf{x}_i) - \Phi (\textbf{x}_j)|, j\in K_i$, where $K_i$ is $k\times k$ neighborhoods around pixel $\textbf{x}_i$.
For appearance information, it may be one of RGB image, optical flow, depth map or thermal image.
To extract more distinctive features from multimodal data, we formulate the texture vector $T^a_i = [t^a_{i, 1}, t^a_{i, 2}, ..., t^a_{i, k^2}]$ as $t^a_{i, j} = exp(-\alpha \sum_m\|\textbf{x}_{i}^{(m)} - \textbf{x}_{j}^{(m)}\|^2), j\in K_i$, where $\alpha$ is a hyperparameter and $\textbf{x}_{i}^{(m)}$ means the appearance data of modal $m$.
It is noteworthy that a small difference between two pixels produces a small element in $T^s$, but a large element in $T^a$.
Therefore, $T^s_i \cdot (T^a_i)^T$ is defined as the matching penalty for considering pixel $\textbf{x}_{i}$ as saliency boundary, contrary to similarity.
Finally, the formula of our $L_{btm}$ is:
\begin{equation}
\label{eqn:acl}
L_{btm} = \frac{\sum_{i}b_iT^s_i \cdot (T^a_i)^T}{\sum_{i}b_i},
\end{equation}
where $b_i$ is the binary boundary mask of saliency prediction.

\begin{figure}[!t]
\centering
\includegraphics[width=0.45 \textwidth]{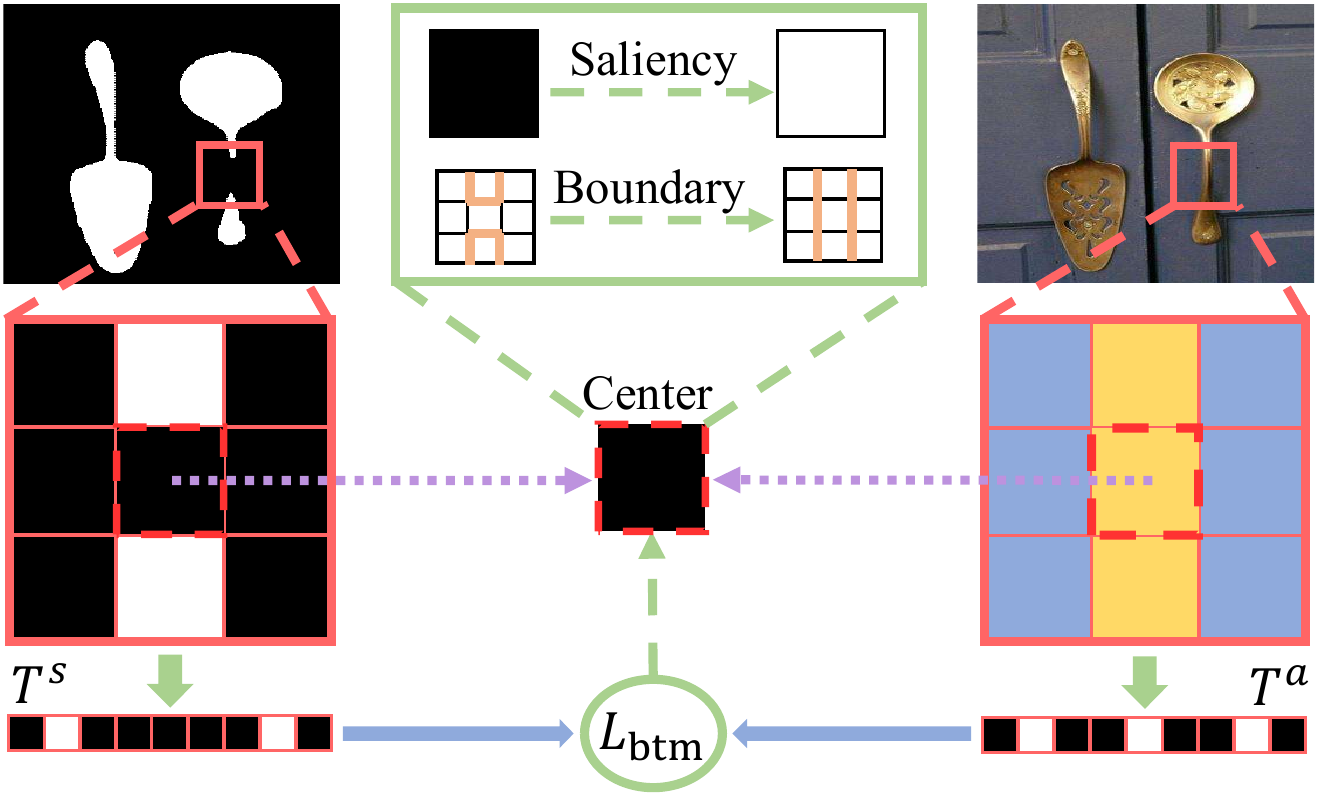}
\vspace{-0.1in}
\caption{Graphical illustration of the proposed $L_{btm}$.}
\label{fig:acloss}
\vspace{-0.15in}
\end{figure}

An intuitive example is shown in Fig. \ref{fig:acloss}.
For the predicted boundary pixels, we find some nearby pixels that differ significantly in saliency scores.
In general, the appearances between boundary and these pixels are also different because they locate within salient objects and backgrounds, respectively.
If not the case, we expect the saliency score of this boundary pixel to be close to those nearby pixels, so that the boundary will shift in the opposite direction.
After an iterative training process, our method can align the predicted saliency boundaries with image edges.

\subsection{Multi-scale Consistency}
Salient objects are consistent in multi-scale inputs.
Thus, we resize input images to a reference scale and encourage our method to produce consistent predictions by:
\begin{equation}
L_{ms} = \sum_i|y_i - resize(\hat y_i)|^2,
\end{equation}
where $y_i$ and $\hat y_i$ are saliency predictions of different scales.


\subsection{Training a Generalized Detector}
Following previous USOD methods \cite{usps,dcfd,a2s}, we generate pseudo labels $\ddot{Y}$ based on saliency predictions $Y$ through $\ddot{Y} = CRF(Y)$ and use them to train an extra saliency detector with the IOU loss.
For RGB SOD, we use the same detector as the previous method \cite{a2s}.
For multimodal SOD tasks, we employ the MIDD \cite{midd} network as our detector.
To ensure a fair comparison with existing SOD methods, we train the extra detector only using task-specific data and corresponding pseudo labels. 

\begin{table*}[t]
\caption{Experiment results on SOD benchmarks.
``Sup.'' indicates the supervised signals used to train SOD methods.
``F'', ``W'' and ``U'' mean fully-supervised, weakly-supervised and unsupervised, respectively.
Best scores are in bold.
}
\label{tab:result}
\centering
\renewcommand\tabcolsep{2pt}
\vspace{-0.1in}
\begin{tabular}{l|c|c|ccc|ccc|ccc|ccc|ccc|ccc}
\hline
\multirow{2}{*}{Methods} & \multirow{2}{*}{Year} & \multirow{2}{*}{Sup.} & \multicolumn{3}{c|}{ECSSD} & \multicolumn{3}{c|}{MSRA-B} & \multicolumn{3}{c|}{DUT-O} & \multicolumn{3}{c|}{PASCAL-S}& \multicolumn{3}{c|}{DUTS-TE}& \multicolumn{3}{c}{HKU-IS} \\\cline{4-21}
& & & $F_{\beta}\negmedspace \uparrow$ & $E_{\xi}\negmedspace \uparrow$ & $\mathcal{M}\negmedspace \downarrow$ & $F_{\beta}\negmedspace \uparrow$ & $E_{\xi}\negmedspace \uparrow$ & $\mathcal{M}\negmedspace \downarrow$ & $F_{\beta}\negmedspace \uparrow$ & $E_{\xi}\negmedspace \uparrow$ & $\mathcal{M}\negmedspace \downarrow$ & $F_{\beta}\negmedspace \uparrow$ & $E_{\xi}\negmedspace \uparrow$ & $\mathcal{M}\negmedspace \downarrow$ & $F_{\beta}\negmedspace \uparrow$ & $E_{\xi}\negmedspace \uparrow$ & $\mathcal{M}\negmedspace \downarrow$ & $F_{\beta}\negmedspace \uparrow$ & $E_{\xi}\negmedspace \uparrow$ & $\mathcal{M}\negmedspace \downarrow$   \\\hline
\multicolumn{21}{c}{Trained on MSRA-B} \\ \hline
SBF \cite{sbf} & 2017 & U & .812 & .878 & .087 & .867 & .929 & .058 & .611 & .771 & .106 & .711 & .795 & .131 & .627 & .785 & .105 & .805 & .895 & .074 \\
MNL \cite{mnl} & 2018 & U & .874 & .906 & .069 & .881 & .932 & .053 & .683 & .821 & .076 & .792 & .846 & \textbf{.091} & -- & -- & -- & .874 & .932 & .047 \\
USPS \cite{usps} & 2019 & U & .875 & .903 & .064 & .896 & .938 & .042 & .715 & .839 & .069 & .770 & .828 & .107 & .730 & .840 & .072 & .880 & .933 & .043 \\
DCFD \cite{dcfd} & 2022 & U & .880 & .900 & .064 & .903 & .938 & .041 & \textbf{.731} & .838 & \textbf{.064} & .773 & .830 & .105 & .744 & .832 & .068 & .887 & .926 & .044 \\
A2S \cite{a2s} & 2022 & U & .888 & .911 & .064 & .902 & .941 & .041 & .719 & .841 & .069 & .790 & .838 & .106 & .750 & .860 & .065 & .887 & .937 & .042 \\
Ours & -- & U & \textbf{.902} & \textbf{.923} & \textbf{.056} & \textbf{.912} & \textbf{.948} & \textbf{.036} & \textbf{.731} & \textbf{.851} & .065 & \textbf{.803} & \textbf{.848} & .099 & \textbf{.767} & \textbf{.871} & \textbf{.061} & \textbf{.891} & \textbf{.939} & \textbf{.041} \\\hline
\multicolumn{21}{c}{Trained on DUTS-TR} \\ \hline
MINet \cite{minet} & 2020 & F & .924 & \textbf{.953} & .033 & .903 & .948 & .038 & .756 & .873 & .055 & .842 & .899 & .064 & .828 & .917 & .037 & .908 & \textbf{.961} & .028 \\
LDF \cite{ldf} & 2020 & F & .930 & .951 & .034 & .902 & .944 & .037 & .773 & .881 & .052 & \textbf{.853} & \textbf{.903} & \textbf{.062} & .855 & .929 & .034 & .914 & .960 & .028 \\
KRN \cite{krn} & 2021 & F & \textbf{.931} & .951 & \textbf{.032} & \textbf{.911} & \textbf{.950} & \textbf{.036} & \textbf{.793} & \textbf{.893} & \textbf{.050} & .851 & .894 & .068 & \textbf{.865} & \textbf{.934} & \textbf{.033} & \textbf{.920} & \textbf{.961} & \textbf{.027} \\\hline
WSSA \cite{wssa} & 2020 & W & .870 & .917 & .059 & .869 & .929 & .049 & .703 & .845 & .068 & .785 & .855 & .096 & .742 & .869 & .062 & .860 & .932 & .047 \\
MFNet \cite{mfnet} & 2021 & W & .844 & .889 & .084 & .872 & .923 & .059 & .621 & .784 & .098 & .756 & .824 & .115 & .693 & .832 & .079 & .839 & .919 & .058 \\
SCW \cite{lsc} & 2021 & W & \textbf{.900} & \textbf{.931} & \textbf{.049} & \textbf{.898} & \textbf{.940} & \textbf{.040} & \textbf{.758} & \textbf{.862} & \textbf{.060} & \textbf{.827} & \textbf{.879} & \textbf{.080} & \textbf{.823} & \textbf{.890} & \textbf{.049} & \textbf{.896} & \textbf{.943} & \textbf{.038} \\ \hline
EDNS \cite{edns} & 2020 & U & .872 & .906 & .068 & .880 & .932 & .051 & .682 & .821 & .076 & .801 & .846 & .097 & .735 & .847 & .065 & .874 & .933 & .046 \\
SelfMask \cite{selfmask} & 2022 & U & .856 & .920 & .058 & .844 & .925 & .050 & .668 & .815 & .078 & .774 & .856 & .087 & .714 & .848 & .063 & .819 & .915 & .053 \\
DCFD \cite{dcfd} & 2022 & U & .888 & .915 & .059 & .888 & .930 & .045 & .710 & .837 & .070 & .795 & .860 & .090 & .764 & .855 & .064 & .889 & .935 & .042 \\
Ours$_{s1}$  & -- & U & .847 & .912 & .057 & .849 & .925 & .054 & .622 & .773 & .111 & .763 & .840 & .093 & .676 & .814 & .082 & .819 & .914 & .053 \\
Ours & -- & U & \textbf{.916} & .938 & .044 & \textbf{.904} & .944 & \textbf{.039} & \textbf{.745} & \textbf{.863} & \textbf{.061} & .830 & .882 & .074 & \textbf{.810} & \textbf{.901} & \textbf{.047} & \textbf{.902} & \textbf{.947} & \textbf{.037} \\
Ours$_{mm}$  & -- & U & .913 & \textbf{.942} & \textbf{.043} & .899 & \textbf{.945} & \textbf{.039} & .737 & .855 & .064 & \textbf{.831} & \textbf{.890} & \textbf{.073} & .803 & .899 & .048 & .898 & .946 & \textbf{.037} \\\hline

\end{tabular}
\end{table*}

\begin{figure*}[t]
\centering
\includegraphics[width=1 \textwidth]{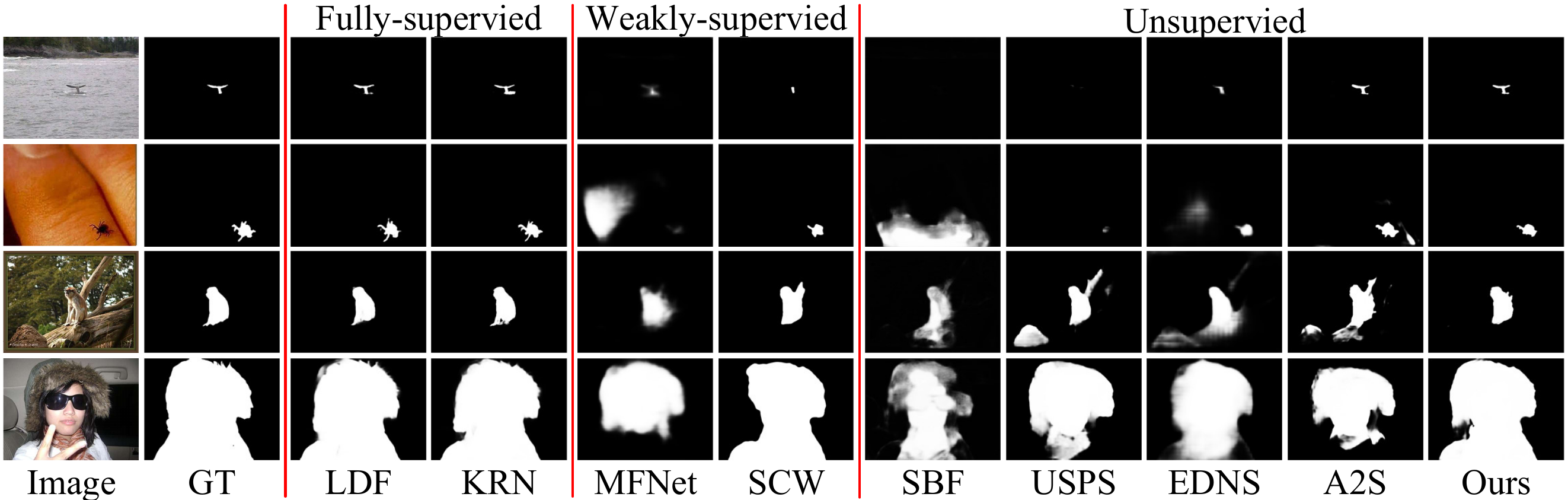}
\vspace{-0.2in}
\caption{Visual comparison between state-of-the-art SOD methods and ours.}
\label{fig:visual}
\vspace{-0.1in}
\end{figure*}

\section{Experiments}
\paragraph{Implementation details.}
All experiments are implemented on a single GTX 1080 Ti GPU.
The batch size is 8 and input images are resized to $320^2$.
The reference scale is randomly selected from $(192^2, 256^2, 384^2, 448^2)$.
Only horizontal flipping is employed as our data augmentation strategy.
We train our method for 20 epochs using the SGD optimizer with an initial learning rate of 0.1, which is decayed linearly.
$\alpha$, $\lambda_c$, $\lambda_b$, and $\lambda_m$ are set to 200, 1, 0.05 and 1, respectively.
We train the extra detectors for 10 epochs using the SGD optimizer with a learning rate of 0.005.

\begin{table*}[t]
\caption{Results on RGB-D SOD benchmarks. ``F'' and ``U'' mean fully-supervised and unsupervised, respectively.
}
\label{tab:rgbd}
\centering
\renewcommand\tabcolsep{7pt}
\small
\vspace{-0.1in}
\begin{tabular}{l|c|c|ccc|ccc|ccc|ccc}
\hline
\multirow{2}{*}{Methods} & \multirow{2}{*}{Year} & \multirow{2}{*}{Sup.} & \multicolumn{3}{c|}{RGBD135} & \multicolumn{3}{c|}{NJUD} & \multicolumn{3}{c|}{NLPR} & \multicolumn{3}{c}{SIP} \\\cline{4-15}
& & & $F_{\beta}\negmedspace \uparrow$ & $E_{\xi}\negmedspace \uparrow$ & $\mathcal{M}\negmedspace \downarrow$ & $F_{\beta}\negmedspace \uparrow$ & $E_{\xi}\negmedspace \uparrow$ & $\mathcal{M}\negmedspace \downarrow$ & $F_{\beta}\negmedspace \uparrow$ & $E_{\xi}\negmedspace \uparrow$ & $\mathcal{M}\negmedspace \downarrow$ & $F_{\beta}\negmedspace \uparrow$ & $E_{\xi}\negmedspace \uparrow$ & $\mathcal{M}\negmedspace \downarrow$   \\\hline
DSA2F \cite{dsa2f} & 2021 & F & .899 & .958 & .021 & .901 & .937 & .039 & .897 & .953 & .024 & -- & -- & --  \\
SPNet \cite{spnet} & 2021 & F & \textbf{.927} & \textbf{.984} & \textbf{.013} & -- & -- & -- & .903 & .959 & \textbf{.021} & \textbf{.893} & \textbf{.931} & \textbf{.043 } \\
CCFE \cite{ccfe} & 2022 & F & .911 & .964 & .020 & \textbf{.914} & \textbf{.953} & \textbf{.032} & \textbf{.907} & \textbf{.962} & \textbf{.021} & .889 & .923 & .047  \\
\hline
DSU \cite{dsu}  & 2022 & U & .767 & .895 & .061 & .719 & .797 & .135 & .745 & .879 & .065 & .619 & .774& .156  \\
Ours & -- & U & .834 & .945 & .037 & .787 & .829 & .109 & .834 & .924 & .043 & .716 & .784 & .124  \\
Ours$_{mm}$ & -- & U & \textbf{.877} & \textbf{.946} & \textbf{.029} & \textbf{.862} & \textbf{.908} & \textbf{.060} & \textbf{.852} & \textbf{.931} & \textbf{.034} & \textbf{.873} & \textbf{.925} & \textbf{.051} \\
\hline
\end{tabular}
\end{table*}

\begin{table*}[t]
\caption{Results on VSOD benchmarks. ``F'', ``W'' and ``U'' mean fully-supervised, weakly-supervised and unsupervised, respectively.
}
\label{tab:vsod}
\centering
\renewcommand\tabcolsep{7pt}
\small
\vspace{-0.1in}
\begin{tabular}{l|c|c|ccc|ccc|ccc|ccc}
\hline
\multirow{2}{*}{Methods} & \multirow{2}{*}{Year} & \multirow{2}{*}{Sup.} & \multicolumn{3}{c|}{DAVSOD} & \multicolumn{3}{c|}{DAVIS} & \multicolumn{3}{c|}{SegV2} & \multicolumn{3}{c}{FBMS} \\\cline{4-15}
& & & $F_{\beta}\negmedspace \uparrow$ & $E_{\xi}\negmedspace \uparrow$ & $\mathcal{M}\negmedspace \downarrow$ & $F_{\beta}\negmedspace \uparrow$ & $E_{\xi}\negmedspace \uparrow$ & $\mathcal{M}\negmedspace \downarrow$ & $F_{\beta}\negmedspace \uparrow$ & $E_{\xi}\negmedspace \uparrow$ & $\mathcal{M}\negmedspace \downarrow$ & $F_{\beta}\negmedspace \uparrow$ & $E_{\xi}\negmedspace \uparrow$ & $\mathcal{M}\negmedspace \downarrow$   \\\hline
PCSA \cite{pcsa} & 2020 & F & .556 & .749 & .077 & .794 & .922 & .022 & .789 & .931 & .019 & .783 & .868 & .042 \\
TENet \cite{tenet}  & 2020 & F & \textbf{.595} & \textbf{.773} & \textbf{.067} & \textbf{.821} & \textbf{.941} & \textbf{.017} & -- & -- & -- & \textbf{.851} & \textbf{.915} & \textbf{.026}  \\
STVS \cite{stvs}  & 2021 & F & .563 & .764 & .080 & .812 & .940 & .022 & \textbf{.835} & \textbf{.950} & \textbf{.016} & .821 & .903 & .042  \\
WSVSOD \cite{wvsod}  & 2021 & W & .492 & .710 & .103 & .731 & .900 & .036 & .711 & .909 & .031 & .736 & .840 & .084  \\
\hline
Ours  & -- & U & .534 & .747 & \textbf{.084} & .751 & \textbf{.913} & .042 & .751 & .914 & .033 & .732 & .794 & .100  \\
Ours$_{mm}$ & -- & U & \textbf{.547} & \textbf{.762} & .085 & \textbf{.756} & .908 & \textbf{.037} & \textbf{.808} & \textbf{.927} & \textbf{.021} & \textbf{.795} & \textbf{.876} & \textbf{.060} \\
\hline
\end{tabular}
\end{table*}

\begin{table*}[!t]
\caption{Results on RGB-T SOD benchmarks. ``F'' and ``U'' mean fully-supervised and unsupervised, respectively.}
\label{tab:rgbt}
\centering
\renewcommand\tabcolsep{10pt}
\small
\vspace{-0.1in}
\begin{tabular}{l|c|c|ccc|ccc|ccc}
\hline
\multirow{2}{*}{Methods} & \multirow{2}{*}{Year} & \multirow{2}{*}{Sup.} & \multicolumn{3}{c|}{VT5000} & \multicolumn{3}{c|}{VT1000} & \multicolumn{3}{c}{VT821} \\\cline{4-12}
& & & $F_{\beta}\negmedspace \uparrow$ & $E_{\xi}\negmedspace \uparrow$ & $\mathcal{M}\negmedspace \downarrow$ & $F_{\beta}\negmedspace \uparrow$ & $E_{\xi}\negmedspace \uparrow$ & $\mathcal{M}\negmedspace \downarrow$ & $F_{\beta}\negmedspace \uparrow$ & $E_{\xi}\negmedspace \uparrow$ & $\mathcal{M}\negmedspace \downarrow$   \\\hline
MIED \cite{mied} & 2020 & F & .761 & .880 & .050 & .853 & .928 & .030 & .760 & .877 & .050  \\
MIDD \cite{midd}  & 2021 & F & .801 & .899 & .043 & .882 & .942 & .027 & .805 & .898 & .045  \\
APNet \cite{apnet} & 2021 & F & .821 & .918 & .035 & .885 & .951 & .021 & .818 & .912 & .034 \\
CCFE \cite{ccfe}  & 2022 & F & \textbf{.859} & \textbf{.937} & \textbf{.030} & \textbf{.906} & \textbf{.963} & \textbf{.018} & \textbf{.857} & \textbf{.934} & \textbf{.027} \\
\hline
Ours  & -- & U & \textbf{.810} & \textbf{.904} & \textbf{.046} & \textbf{.885} & \textbf{.939} & \textbf{.031} & \textbf{.805} & \textbf{.900} & \textbf{.043}   \\
Ours$_{mm}$ & -- & U & .807 & .903 & .047 & .881 & \textbf{.939} & .032 & \textbf{.805} & .899 & .044 \\
\hline
\end{tabular}
\vspace{-0.1in}
\end{table*}

\textbf{Datasets.}
In our experiments, we follow the prevalent settings of different SOD tasks.
Specifically, for RGB SOD, we use the training subsets of MSRA-B \cite{msra} or DUTS \cite{duts} to train our method, respectively.
ECSSD \cite{ecssd}, PASCAL-S \cite{pascal-S}, HKU-IS \cite{hku-is}, DUTS-TE \cite{duts}, DUT-O \cite{DUT-OMRON}, and the testing subset of MSRA-B are employed for evaluation.
For RGB-D SOD, we choose 2185 samples from NLPR \cite{nlpr} and NJUD \cite{njud} as the training set.
RGBD135 \cite{rgbd135}, SIP \cite{sip} and the testing subsets of NJUD and NLPR are employed for evaluation.
For RGB-T SOD, 2500 images in VT5000 \cite{vt5000} are for training, while VT1000 \cite{vt1000}, VT821 \cite{vt821} and the rest 2500 images in VT5000 are for testing.
For video SOD, we choose the training splits of DAVIS \cite{davis} and DAVSOD \cite{ssav} to train our method.
Moreover, we randomly select 5 frames of each video in DAVSOD to avoid overfitting.
The testing splits of DAVIS, DAVSOD, SegV2 \cite{segv2} and FBMS \cite{fbms} are employed for evaluation.


\textbf{Metrics.}
We adopt three criteria for evaluation, including ave-$F_{\beta}$, Mean Absolute Error ($\mathcal{M}$) and E-Measure ($E_{\xi}$) \cite{Emeasure}.
Specifically, 
$
F_{\beta} = \frac{(1+\beta^{2})\times Precision \times Recall}{\beta^{2}\times Precision + Recall},
$
where $\beta^{2}$ is set to 0.3 \cite{THUR15K}.
The ave-$F_{\beta}$ is the $F_{\beta}$ scores by setting the threshold as two times mean values.
$\mathcal{M}$ is the absolute error between predictions and ground truth. 
$E_{\xi}$ measures the global statistics and local pixel matching information.


\subsection{Results on RGB SOD}
As shown in Tab. \ref{tab:result}, we compare the proposed method with \textbf{fully-supervised} methods, MINet \cite{minet}, LDF \cite{ldf}, and KRN \cite{krn}, \textbf{weakly-supervised} methods, WSSA \cite{wssa}, MFNet \cite{mfnet}, and SCW \cite{lsc}, and \textbf{unsupervised} methods, SBF \cite{sbf}, MNL \cite{mnl}, USPS \cite{usps}, EDNS \cite{edns}, A2S \cite{a2s}, DCFD \cite{dcfd}, and SelfMask \cite{selfmask}.
We list our results with different settings: (1) saliency results without extra detector and post-processing, denoted as ``Ours$_{s1}$''; (2) our full method as ``Ours''; (3) training our method using multimodal data, while only DUTS-TR or MSRA-B datasets with pseudo labels are employed to train the extra detector (``Ours$_{mm}$'').  

Either training on MSRA-B or DUTS-TR, the proposed method achieves significant improvements compared to existing USOD methods.
Moreover, our unsupervised method is competitive to recent weakly-supervised SCW \cite{lsc} and fully-supervised methods KRN \cite{krn}.
Furthermore, the results of ``Ours$_{s1}$'' prove that our method extracts precise saliency knowledge from training samples, but somehow weaken when deploying on unseen images. 
In addition, the similar results of ``Ours$_{mm}$'' and ``Ours'' imply that extra multimodal data may not bring significant improvements to RGB SOD task.
As proved in \cite{wang_survey}, the large-scale DUTS-TR dataset is saturated for training saliency detectors, even in the unsupervised case.

A qualitative comparison is illustrated in Fig. \ref{fig:visual}.
Overall, our saliency predictions are much more precise than other unsupervised methods.
For example, in the first example, our method and two fully-supervised methods well segment target object with precise boundaries, while other methods fail to capture the tiny salient object.

\subsection{Results on Multimodal SOD}
We conduct more experiments on multimodal tasks, including RGB-D, RGB-T and video SOD.
Noted that we pre-compute the optical flow for each video frame as an extra modality.
To verify the improvements brought by multimodal data, we build two sets for training:
(1) Task-specific data (``Ours'').
For example, for RGB-T SOD task, we only use 2500 RGB-thermal image pairs for training as prevalent RGB-T SOD methods \cite{mied,midd,apnet};
(2) Multimodal data from the training sets of four SOD tasks (``Ours$_{mm}$'').

Overall, our method reports state-of-the-art performance on all multimodal SOD tasks.
Specifically, for the RGB-D SOD task in Tab. \ref{tab:rgbd}, our method surpasses the latest unsupervised method DSU \cite{dsu}.
For video and RGB-T SOD tasks in Tab. \ref{tab:vsod} and \ref{tab:rgbt}, to the best of our knowledge, our method is the first unsupervised method and is competitive to supervised methods.
Furthermore, our method trained on multimodal data (``Ours$_{mm}$'') produce high-quality pseudo labels for all multimodal SOD datasets simultaneously and has achieved performance improvements on most tasks.

\begin{table}[t]
\caption{Label quality on multimodal SOD datasets.
}
\label{tab:mm}
\centering
\renewcommand\tabcolsep{4pt}
\small
\vspace{-0.1in}
\begin{tabular}{l|cc|cc|cc|cc}
\hline
\multirow{2}{*}{Methods} & \multicolumn{2}{c|}{RGB} & \multicolumn{2}{c|}{RGB-D} & \multicolumn{2}{c|}{VSOD} & \multicolumn{2}{c}{RGB-T} \\\cline{2-9}
& $F_{\beta}\negmedspace \uparrow$  & $\mathcal{M}\negmedspace \downarrow$ & $F_{\beta}\negmedspace \uparrow$  & $\mathcal{M}\negmedspace \downarrow$ & $F_{\beta}\negmedspace \uparrow$ &  $\mathcal{M}\negmedspace \downarrow$ & $F_{\beta}\negmedspace \uparrow$  & $\mathcal{M}\negmedspace \downarrow$   \\\hline
Ours         & .917 & .038 & .746 & .110 & .557 & .099 & .932 & .029  \\
Ours$_{rgb}$ & .917 & .040 & .804 & .068 & .613 & .089 & .920 & .030 \\
Ours$_{mm}$  & .918 & .040 & .826 & .062 & .637 & .079 & .923 & .029  \\
\hline
\end{tabular}
\vspace{-0.1in}
\end{table}

\subsection{Ablation Study}
\begin{figure}[!t]
\centering
\includegraphics[width=0.48 \textwidth]{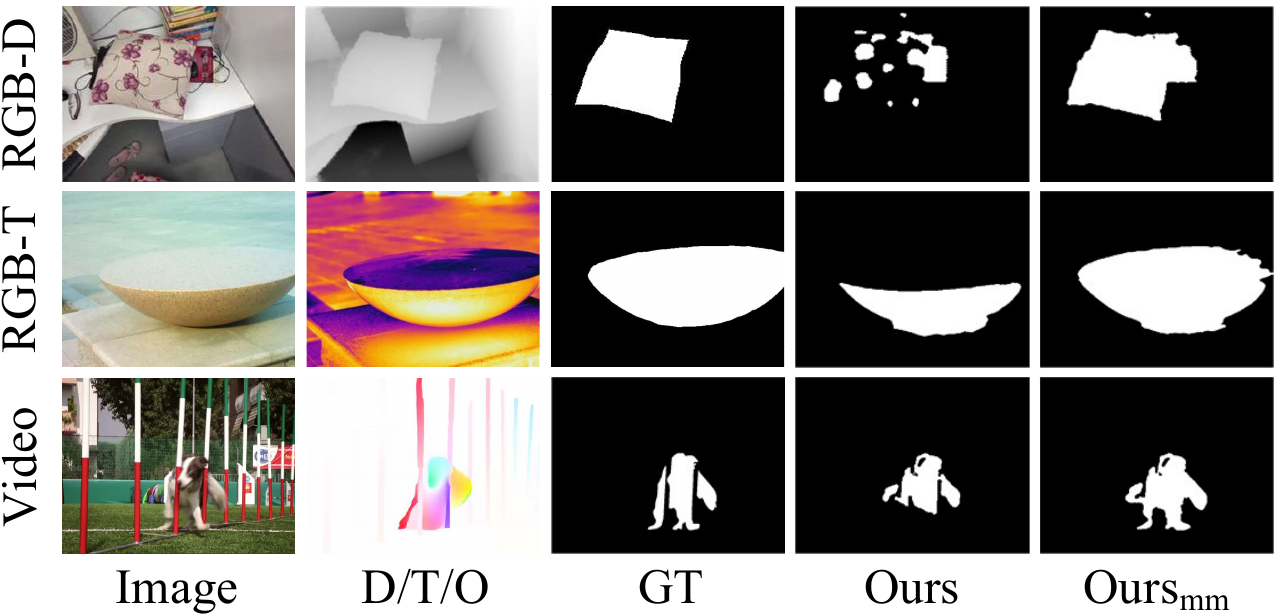}
\caption{Examples of the generated pseudo labels. The ``D/T/O'' means depth map, thermal image or optical flow.}
\label{fig:mm_show}
\end{figure}

\paragraph{Label quality comparison.}
In Tab. \ref{tab:mm}, we exhibit the scores of pseudo labels on multimodal SOD datasets.
Except for the above two training sets, we employ an additional set that collects all RGB images from four tasks to train our method, denoted as ``Ours$_{rgb}$''.
Overall, ``Ours$_{mm}$'' reports more generalized results on all datasets.
Moreover, training on task-specific data (``Ours'') slightly surpasses the performance of ``Ours$_{mm}$'' on the training sets of RGB and RGB-T tasks.
Since our method is trained without ground truth, fitting a hybrid dataset may cause a slight performance drop on some subsets.
In addition, the comparison between ``Ours$_{rgb}$'' and ``Ours$_{mm}$'' prove that multimodal data significantly improves the quality of the generated pseudo labels.
For an intuitive comparison, we show some examples in Fig. \ref{fig:mm_show}.
Trained with task-specific data, our method precisely localizes salient regions but fails to segment the whole objects.
With more training data, the network perceives fine-grained concepts of target objects, resulting in more complete segmentations.
 

\textbf{Impact of initial saliency cues.}
Our method is based on the activation maps of a pre-trained network \cite{resnet,mocov2} instead of traditional methods \cite{mc,rbd,dsr} as other DL-based USOD methods \cite{sbf,mnl,usps,msum,dcfd}.
We compare the performance of these two types of designs in Tab. \ref{tab:saliency_cue}.
Overall, the activation map is worse before training, whereas better than traditional methods \cite{mc,dsr,rbd} after training.
The activation maps are dynamic during training, such that the network strengthens the learned saliency knowledge incrementally.
On the contrary, using traditional methods, the network learns from fixed labels throughout the training process, and thus fits the biased knowledge in those traditional methods.
In addition, traditional methods introduce extra computation loads, while the pre-trained network is indispensable for all DL-based USOD methods. 

\begin{table}[t]
\caption{Comparison between different saliency cues.}
\renewcommand\tabcolsep{4.5pt}
\label{tab:saliency_cue}
\vspace{-0.1in}
\begin{tabular}{l|ccc|ccc}
\hline
\multirow{2}{*}{Saliency cues} & \multicolumn{3}{c|}{Before training}& \multicolumn{3}{c}{After training}  \\\cline{2-7}
  & $F_{\beta} \uparrow$ & $E_{\xi} \uparrow$ & $\mathcal{M} \downarrow$ & $F_{\beta} \uparrow$ & $E_{\xi} \uparrow$ & $\mathcal{M} \downarrow$ \\\hline
 MC \cite{mc} & .810 & .877 & .145 & .889 & .914 & .053 \\
 DSR \cite{dsr} & .780 & .867 & .118 & .886 & .901 & .059 \\
 RBD \cite{rbd} & .803 & .883 & .109 & .891 & .912 & .053 \\\hline
 Ours & .451 & .659 & .353 & .914 & .940 & .039 \\\hline
\end{tabular}
\vspace{-0.1in}
\end{table}

\textbf{Effectiveness of loss functions.}
We conduct ablation studies on various combinations of loss functions in Tab. \ref{tab:ablation} and more details are introduced as follows.

In ablation study A, our method receives continuous improvements by appending new losses ($L_{ms}$ and $L_{btm}$) or replacing the original $L_{adb}$ to our $L_{csd}$.
This experiment prove that these three losses can assist the network mine more detailed saliency knowledge from different perspectives.
More importantly, these losses are supplementary to each other, such that their combination can collaborate to produce high-quality pseudo labels for SOD datasets.

In ablation study B, our $L_{csd}$ loss can provide precise saliency localization information and thus achieves the best performance among all competitors.
Specifically, $L_{bce}$ and $L_1$ fail to mine detailed and reliable saliency knowledge from noisy saliency cues.
$L_{adb}$ \cite{a2s} improves the robustness of the learned saliency knowledge by focusing on easy samples, and thus surpasses the $L_1$ loss.
$L_{spl}$ \cite{SPL} also employs a dynamic schedule, but its binary weights filter out latent saliency knowledge in hard samples.

\begin{table}[t]
\caption{Ablation studies on loss functions.}
\renewcommand\tabcolsep{5pt}
\label{tab:ablation}
\vspace{-0.1in}
\begin{tabular}{l|l|ccc}
\hline
Tag & Loss  & $F_{\beta} \uparrow$ & $E_{\xi} \uparrow$ & $\mathcal{M} \downarrow$ \\\hline
A1 & $L_{adb}$ \cite{a2s} (Baseline) & .882 & .915 & .071  \\
A2 & $L_{adb}$ \cite{a2s} + $L_{ms}$ & .891 & .921 & .066  \\
A3 & $L_{csd}$ + $L_{ms}$ & .908 & .937 & .050  \\
A4 & $L_{csd}$ + $L_{btm}$ + $L_{ms}$ (Ours) & \textbf{.917} & \textbf{.945} & \textbf{.038} \\\hline
B1 & $L_{bce}$ + $L_{btm}$ + $L_{ms}$ & .558 & .540 & .208  \\
B2 & $L_{1}$ + $L_{btm}$ + $L_{ms}$ & .895 & .928 & .044  \\
B3 & $L_{adb}$ \cite{a2s} + $L_{btm}$ + $L_{ms}$  & .903 & .934 & .042  \\
B4 & $L_{spl}$ \cite{SPL} + $L_{btm}$ + $L_{ms}$  & .910 & .938 & .041  \\
B5 & $L_{csd}$ + $L_{btm}$ + $L_{ms}$ (Ours)  & \textbf{.917} & \textbf{.945} & \textbf{.038} \\\hline
C1 & $L_{csd}$ + $L_{c1}$ + $L_{ms}$ & .735 & .819 & .120  \\
C2 & $L_{csd}$ + $L_{lsc}$ \cite{lsc} + $L_{ms}$ & .906 & .923 & .052 \\
C3 & $L_{csd}$ + $L_{c2}$ + $L_{ms}$ & .907 & .926 & .051 \\
C4 & $L_{csd}$ + $L_{btm}$ + $L_{ms}$ (Ours) & \textbf{.917} & \textbf{.945} & \textbf{.038} \\\hline
\end{tabular}
\end{table}


In ablation study C, we compare our $L_{btm}$ with $L_{lsc}$ \cite{lsc} and two variants: (1) using L1 distance for the texture features in appearance space, denoted as $L_{c1}$; (2) removing the boundary masks in $L_{btm}$, denoted as $L_{c2}$.
In general, our $L_{btm}$ outperforms other variants.
Specifically, the L1 distance in $L_{c1}$ causes the texture features to be not distinctive enough, resulting in ambiguous saliency boundaries.
$L_{lsc}$ focuses more on the adjacent pixels so that the boundaries are more susceptible to a limited region rather than a larger patch.
Moreover, using $L_{c2}$, edges within objects or backgrounds may corrupt the learned saliency knowledge.


\begin{table}[t]
\begin{center}
\caption{Label quality on DUTS-TR dataset.}
\vspace{-0.1in}
\renewcommand\tabcolsep{9pt}
\label{tab:discuss}
\begin{tabular}{l|ccc}
\hline
Pre-training & $F_{\beta} \uparrow$ & $E_{\xi} \uparrow$ & $\mathcal{M} \downarrow$ \\\hline
Supervised & .915 & .942 & .039  \\
Unsupervised (Ours) & .917 & .945 & .038  \\\hline
\end{tabular}
\end{center}
\vspace{-0.15in}
\end{table}

\textbf{Supervised or unsupervised pre-training?}
Under the unsupervised setting, whether we can use an encoder with supervised pre-training is controversial.
For the ImageNet dataset \cite{imagenet}, the class labels of images usually indicate the category of the most salient object.
It means that the supervised encoder receives extra saliency knowledge from these manual labels.
Thus, for completely unsupervised SOD, we employ unsupervised MoCo-v2 to initialize our encoder.
As listed in Tab. \ref{tab:discuss}, the performance of our method when using supervised encoder is comparable.


\section{Conclusion}
In this paper, we propose an Unsupervised Salient Object Detection (USOD) method guided by two novel mechanisms. First, we propose a Confidence-aware Saliency Distilling (CSD) to learn saliency knowledge from easy samples to hard ones with noisy labels progressively. Second, we propose a Boundary-aware Texture Matching (BTM) to make the location of saliency prediction more accurate. As a result, the proposed method produces high-quality pseudo labels to train saliency detectors. Experiments on RGB, RGB-D, RGB-T, and video SOD benchmarks prove that our method outperforms existing USOD methods.

{\small
\bibliographystyle{ieee_fullname}
\bibliography{PaperForReview}
}

\clearpage

\begin{appendix}
\twocolumn[{%
\renewcommand\twocolumn[1][]{#1}%
\maketitle
\begin{center}
    \centering
\renewcommand\tabcolsep{11pt}
\captionof{table}{Setting comparison between USOD methods.
``F'' and ``U'' indicate fully-supervised and unsupervised pre-training.
``IN'' and ``CS'' are ImageNet \cite{imagenet} and CityScape \cite{cityscape} datasets, respectively.
}
\begin{tabular}{c|ccccccc}
\hline
Method & Training set & Input & Encoder & Pre-train & Saliency cues & Train time \\ \hline
EDNS \cite{edns} & DUTS-TR & $352\times 352$ & VGG-16 & F-IN & \cite{rbd, mr, gs} & $\textgreater$8h \\
DCFD \cite{dcfd} & DUTS-TR & -- & ResNet-50 & F-IN & \cite{dsr} & -- \\
Ours & DUTS-TR & $320\times 320$ & ResNet-50 & U-IN & No & 4.5h \\\hline
SBF \cite{sbf} & MSRA-B  & $224\times 224$& VGG-16 & F-IN & \cite{mb+, bms, cssd} & $\textgreater$3h \\
MNL \cite{mnl} & MSRA-B  & $425\times 425$& ResNet-101 & F-IN & \cite{rbd, dsr, mc, cssd} & $\textgreater$4h \\
USPS \cite{usps} & MSRA-B  & $432\times 432$ & ResNet-101 & F-CS & \cite{rbd, dsr, mc, cssd} & $\textgreater$30h \\
DCFD \cite{dcfd} & MSRA-B & -- & ResNet-101 & F-CS & \cite{dsr} & -- \\
A2S \cite{a2s} & MSRA-B  & $320\times 320$ & ResNet-50 & U-IN & No & 1h \\
Ours & MSRA-B & $320\times 320$ & ResNet-50 & U-IN & No & 1.3h \\
\hline
\end{tabular}
\label{tab:compare}
\end{center}%
}]

\section{Setup of USOD methods.}
As listed in Tab. \ref{tab:compare}, our method achieves better performance under disadvantage settings.
Specifically, the $320^2$ input of our method is small than most USOD methods, such as $425^2$ for MNL \cite{mnl} and $432^2$ for USPS \cite{usps}.
Moreover, we use ResNet-50 \cite{resnet} as backbone, which is a weakened version of ResNet-101 used in many USOD methods \cite{mnl,usps,dcfd}.
As for pre-training, most existing methods employed the encoders pre-trained with manual annotations of some close-related datasets, such as ImageNet \cite{imagenet} for object recognition and Cityscape \cite{cityscape} for semantic segmentation. 
Such setting indicates that they benefit from the semantic knowledge of manual annotations, which violates the semantic-agnostic definition of the SOD task.
On the contrary, the encoder of our method is pre-trained without using any human annotation.
It means that no semantic knowledge is involved in the whole training process, which accords with the semantic-agnostic definition.
Last, even excluding the additional time of existing methods \cite{edns,sbf,mnl,usps} to extract salience cues using traditional methods, the training time of our method is much less than that of most previous methods.

\begin{figure}
\begin{minipage}{1 \textwidth}
\begin{rotate}{90}
\begin{minipage}{0.1 \textwidth} \centering \scriptsize Image \vspace{0.1in} \end{minipage}
\end{rotate}
\includegraphics[width=1.05in,height=0.7in]{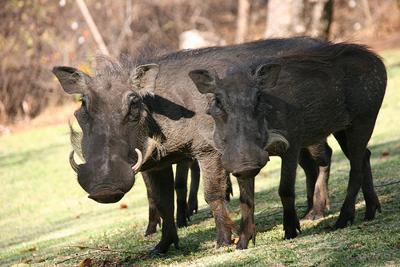}
\includegraphics[width=1.05in,height=0.7in]{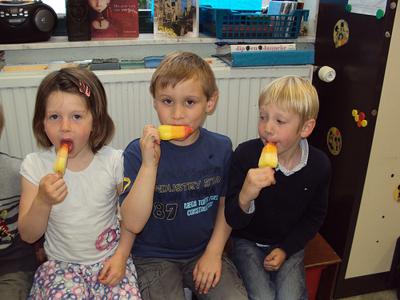}
\includegraphics[width=1.05in,height=0.7in]{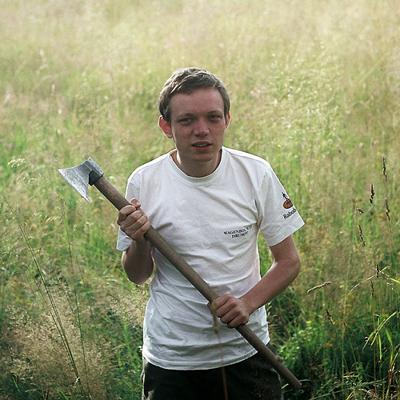}
\end{minipage}

\begin{minipage}{1 \textwidth}
\begin{rotate}{90}
\begin{minipage}{0.1 \textwidth} \centering \scriptsize GT \vspace{0.1in} \end{minipage}
\end{rotate}
\includegraphics[width=1.05in,height=0.7in]{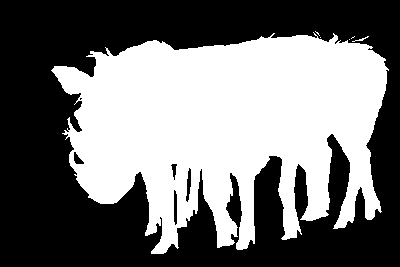}
\includegraphics[width=1.05in,height=0.7in]{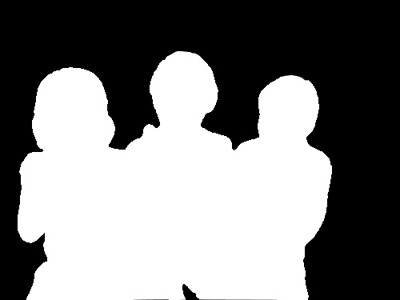}
\includegraphics[width=1.05in,height=0.7in]{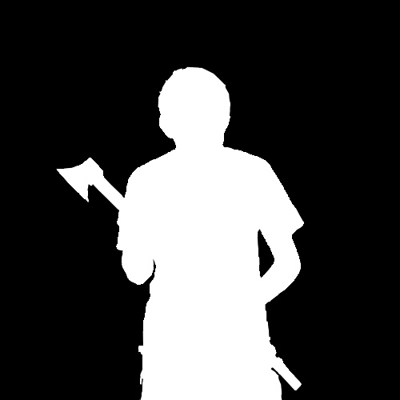}
\end{minipage}

\begin{minipage}{1 \textwidth}
\begin{rotate}{90}
\begin{minipage}{0.1 \textwidth} \centering \scriptsize A1 \vspace{0.1in} \end{minipage}
\end{rotate}
\includegraphics[width=1.05in,height=0.7in]{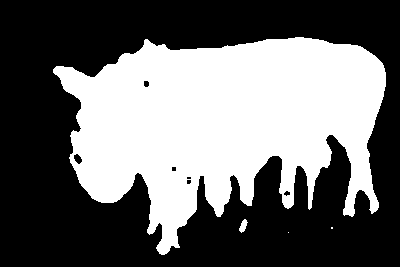}
\includegraphics[width=1.05in,height=0.7in]{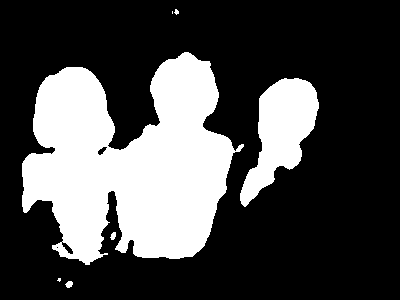}
\includegraphics[width=1.05in,height=0.7in]{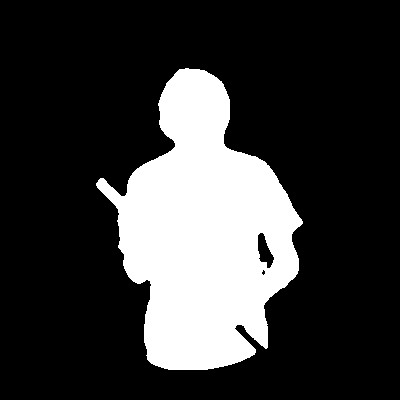}
\end{minipage}

\begin{minipage}{1 \textwidth}
\begin{rotate}{90}
\begin{minipage}{0.1 \textwidth} \centering \scriptsize A2 \vspace{0.1in} \end{minipage}
\end{rotate}
\includegraphics[width=1.05in,height=0.7in]{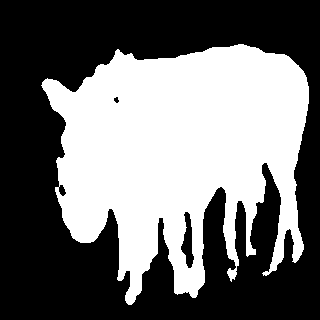}
\includegraphics[width=1.05in,height=0.7in]{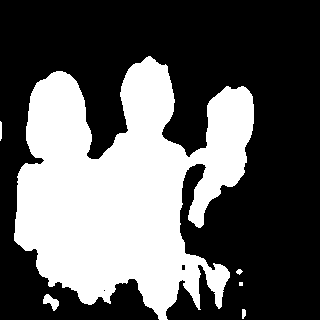}
\includegraphics[width=1.05in,height=0.7in]{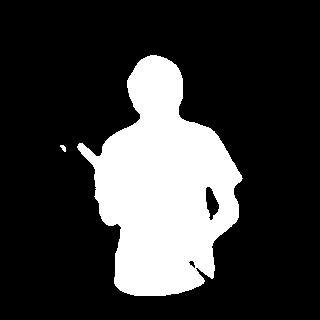}
\end{minipage}

\begin{minipage}{1 \textwidth}
\begin{rotate}{90}
\begin{minipage}{0.1 \textwidth} \centering \scriptsize A3 \vspace{0.1in} \end{minipage}
\end{rotate}
\includegraphics[width=1.05in,height=0.7in]{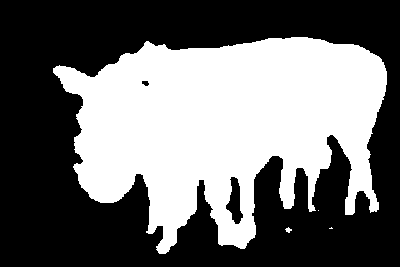}
\includegraphics[width=1.05in,height=0.7in]{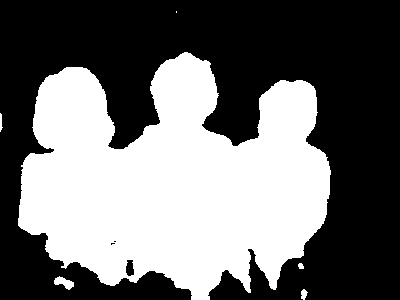}
\includegraphics[width=1.05in,height=0.7in]{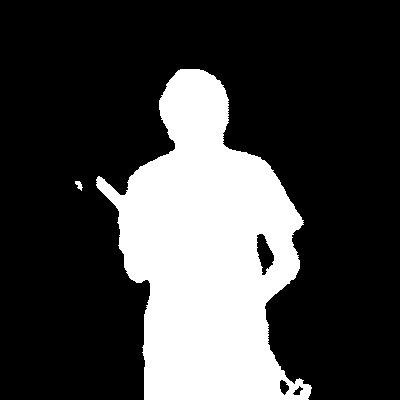}
\end{minipage}

\begin{minipage}{1 \textwidth}
\begin{rotate}{90}
\begin{minipage}{0.1 \textwidth} \centering \scriptsize A4 \vspace{0.1in} \end{minipage}
\end{rotate}
\includegraphics[width=1.05in,height=0.7in]{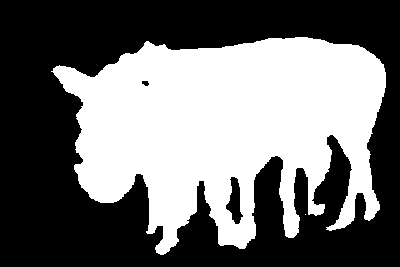}
\includegraphics[width=1.05in,height=0.7in]{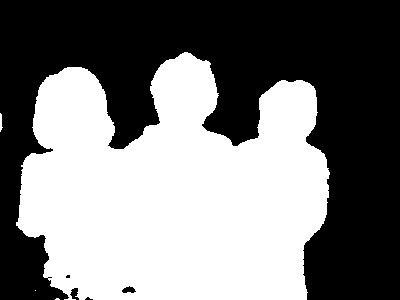}
\includegraphics[width=1.05in,height=0.7in]{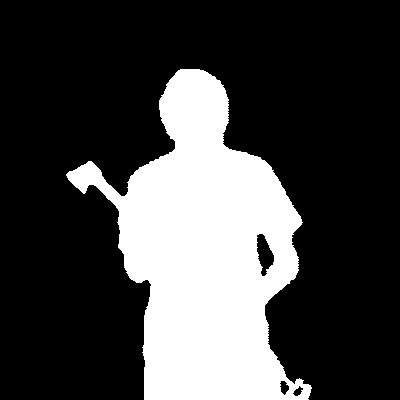}
\end{minipage}

\caption{Saliency predictions of our method with different losses.}
\label{fig:pseudo_show}
\vspace{-0.1in}
\end{figure}

\section{Qualitative comparison of different loss.}
In our manuscript, we exhibit the quantitative results of different losses in ablation study A.
Here, we provide a qualitative comparison in Fig. \ref{fig:pseudo_show}.
Our baseline A1 can accurately localize salient objects in images, but loses many details.
Trained using the proposed losses, the network mines more detailed saliency knowledge progressively and thus precisely predicts the saliency boundaries.

\begin{figure}[!t]
\begin{minipage}{1 \textwidth}
\includegraphics[width=.78in,height=0.57in]{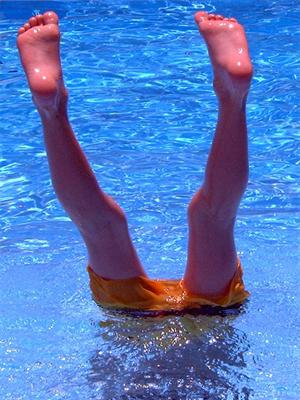}
\includegraphics[width=.78in,height=0.57in]{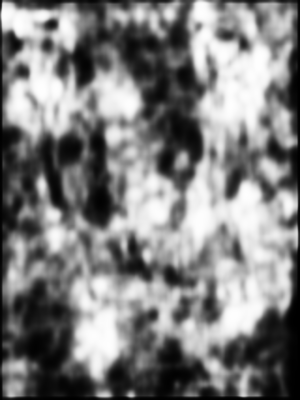}
\includegraphics[width=.78in,height=0.57in]{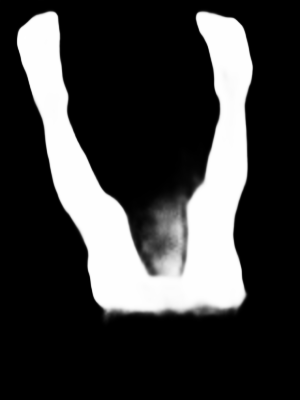}
\includegraphics[width=.78in,height=0.57in]{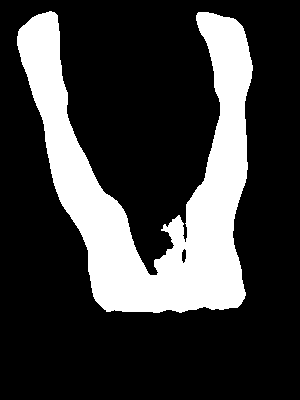}
\end{minipage}

\begin{minipage}{1 \textwidth}
\begin{minipage}{.78in} \centering Image \end{minipage}
\begin{minipage}{.78in} \centering Initial \end{minipage}
\begin{minipage}{.78in} \centering Iter 5 \end{minipage}
\begin{minipage}{.78in} \centering Iter 10 \end{minipage}
\end{minipage}

\begin{minipage}{1 \textwidth}
\includegraphics[width=.78in,height=0.57in]{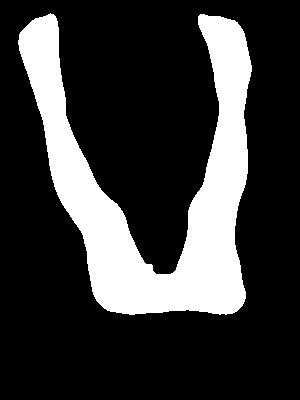}
\includegraphics[width=.78in,height=0.57in]{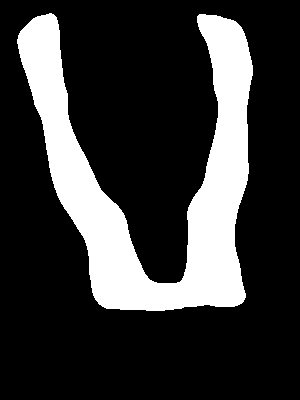}
\includegraphics[width=.78in,height=0.57in]{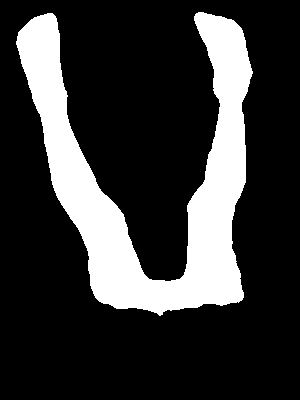}
\includegraphics[width=.78in,height=0.57in]{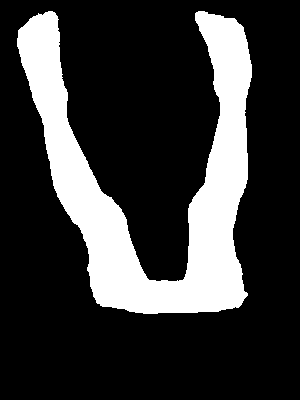}
\end{minipage}

\begin{minipage}{1 \textwidth}
\begin{minipage}{.78in} \centering Iter 15 \end{minipage}
\begin{minipage}{.78in} \centering Iter 20 \end{minipage}
\begin{minipage}{.78in} \centering After CRF \end{minipage}
\begin{minipage}{.78in} \centering GT \end{minipage}
\end{minipage}

\caption{The learned saliency maps during training.}
\label{fig:iteration}
\end{figure}

\section{Visualization of the learned saliency.}
In, Fig. \ref{fig:iteration}, we visualize the learned saliency maps in our method during the training process. 
In the initial stage, our method is able to precisely localize the salient object based on the initial saliency cues, however, some small patches may still be misclassified.
After subsequent tuning process, our method can learn more precise saliency knowledge and thus produce a high-quality pseudo label.

\section{Effect of hyperparameters.}
The performance of our framework is affected by several hyperparameters, including $\alpha$, $\lambda_c$, $\lambda_b$, $\lambda_m$, and $k$. 
We vary these hyperparameters and exhibit their results in Tab. \ref{tab:param}.
The results prove that the values $\alpha=200$, $\lambda_c=1$, $\lambda_b=0.05$ and $\lambda_m=1$ work best in practice.
Our framework is robust to $\alpha \in [100, 300]$, and reports the best performance for $\alpha = 200$.
Moreover, our framework achieves comparable performance for various $\lambda_c$ and $\lambda_m$ values within $[0.5, 1.5]$. 
Furthermore, our framework is sensitive to $\lambda_b$.
Although we observe robust performance for $\lambda_b \in [0.03, 0.07]$, $\lambda_b$ outside this range (\textit{e.g.}, $\lambda_b = 0.01$) seems to induce significant performance drops.

\begin{table}[t]
\begin{center}
\caption{Effect of different hyperparameters.}
\vspace{-0.1in}
\renewcommand\tabcolsep{7pt}
\label{tab:param}
\begin{tabular}{c|c|ccc}
\hline
Parameter & Value & $F_{\beta} \uparrow$ & $E_{\xi} \uparrow$ & $\mathcal{M} \downarrow$ \\\hline
            & 100  & .911 & .932 & .046       \\
            & 150  & .914 & .937 & .043       \\
$\alpha$    & 200  & .917 & .945 & .038       \\
            & 250  & .915 & .942 & .039       \\
            & 300  & .914 & .943 & .039       \\\hline
            & 0.5  & .911 & .937 & .042       \\
            & 0.7  & .915 & .942 & .039       \\
$\lambda_c$ & 1    & .917 & .945 & .038       \\
            & 1.2  & .914 & .943 & .038       \\
            & 1.5  & .913 & .941 & .039       \\\hline
            & 0.01 & .869 & .915 & .054       \\
            & 0.03 & .910 & .938 & .042       \\
$\lambda_b$ & 0.05 & .917 & .945 & .038       \\
            & 0.07 & .914 & .945 & .039       \\
            & 0.09 & .908 & .942 & .040       \\\hline
            & 0.5  & .915 & .943 & .038       \\
            & 0.75 & .915 & .945 & .038       \\
$\lambda_m$ & 1    & .917 & .945 & .038       \\
            & 1.25 & .915 & .943 & .039       \\
            & 1.5  & .914 & .941 & .039       \\\hline
            &    3 & .913 & .937 & .042  \\
k           &    5 & .917 & .945 & .038  \\
            &    7 & .914 & .943 & .039  \\\hline
\end{tabular}
\end{center}
\vspace{-0.2in}
\end{table}

\section{Loss for training extra saliency detectors.}
For fully-supervised SOD methods, there are many choices for the loss functions, such as BCE loss \cite{dss,amulet}, BCE+IOU loss \cite{egnet,minet}, CTLoss \cite{itsd,contour}, BIS(BCE+IOU+SSIM) loss \cite{basnet}.
We employ these losses to train our saliency detector with the generated pseudo labels, as exhibited in Tab. \ref{tab:loss_s2}.
In summary, training our detector with IOU loss achieves the best results compared to other losses.
BCE and CTLoss provide pixel-wise supervised signals, which means that training with these losses is easy to overfit the noises and thus degrade the generalization ability of our detector.
Similarly, SSIM is based on regional statistics and thus is sensitive to noisy regions in pseudo labels.
Unlike the above losses, IOU is robust to pixel-level or region-level noises because it is based on global statistics of saliency predictions.

\begin{table}[t]
\begin{center}
\caption{Different losses for the second stage. }
\vspace{-0.1in}
\renewcommand\tabcolsep{5pt}
\label{tab:loss_s2}
\begin{tabular}{c|ccc|ccc}
\hline
\multirow{2}{*}{Loss} & \multicolumn{3}{c|}{DUT-OMRON} & \multicolumn{3}{c}{ECSSD}  \\\cline{2-7}
 & $F_{\beta} \uparrow$ & $E_{\xi} \uparrow$ & $\mathcal{M} \downarrow$ & $F_{\beta} \uparrow$ & $E_{\xi} \uparrow$ & $\mathcal{M} \downarrow$  \\\hline
BCE & .708 & .834 & .066 & .891 & .924 & .047 \\
BCE+IOU & .726 & .846 & .065 & .894 & .923 & .048 \\
BIS & .716 & .838 & .067 & .886 & .919 & .049 \\
CTLoss & .743 & .862 & .061 & .907 & .914 & .057 \\\hline
IOU & .745 & .863 & .061 & .916 & .938 & .044 \\\hline
\end{tabular}
\end{center}
\end{table}

\begin{table}[t]
\begin{center}
\caption{Performance of SOD methods on X-ray images.}
\vspace{-0.1in}
\renewcommand\tabcolsep{7pt}
\label{tab:xray}
\begin{tabular}{c|c|ccc}
\hline
& Training set & $F_{\beta} \uparrow$ & $E_{\xi} \uparrow$ & $\mathcal{M} \downarrow$ \\\hline
LDF \cite{ldf} & DUTS-TR & .296 & .508 & .315  \\
Ours & DUTS-TR & .530 & .664 & .309  \\
Ours* & DUTS-TR+X-ray & .924 & .943 & .056  \\\hline
\end{tabular}
\end{center}
\end{table}

\begin{figure}[!t]
\begin{minipage}{1 \textwidth}
\includegraphics[width=.63in,height=0.42in]{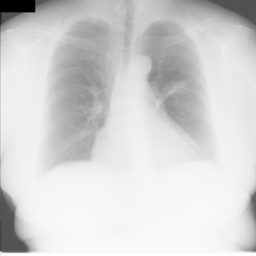}
\includegraphics[width=.63in,height=0.42in]{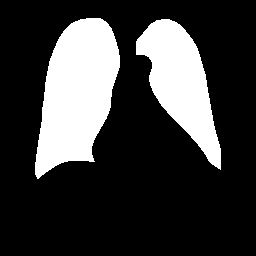}
\includegraphics[width=.63in,height=0.42in]{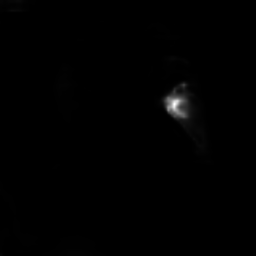}
\includegraphics[width=.63in,height=0.42in]{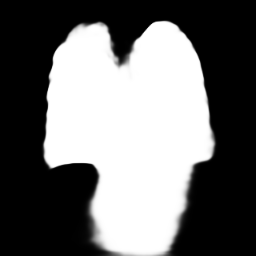}
\includegraphics[width=.63in,height=0.42in]{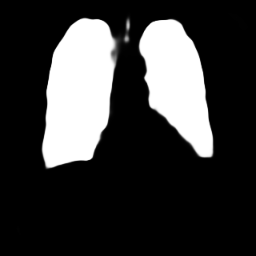}
\end{minipage}

\begin{minipage}{1 \textwidth}
\begin{minipage}{.63in} \centering Image \end{minipage}
\begin{minipage}{.63in} \centering GT \end{minipage}
\begin{minipage}{.63in} \centering LDF \end{minipage}
\begin{minipage}{.63in} \centering Ours \end{minipage}
\begin{minipage}{.63in} \centering Ours* \end{minipage}
\end{minipage}

\vspace{-0.1in}
\caption{Examples of the predicted saliency maps.}
\vspace{-0.1in}
\label{fig:visual}
\end{figure}

\section{Necessity of unsupervised SOD.}
Ideally, a supervised class-agnostic SOD model can handle all scenarios, whereas is hard to obtain in practical. First, SOD methods trained on datasets with limited classes may not perform well on unseen classes, even if class labels are not used during training. Second, SOD methods trained on a certain style of images (e.g., natural images) do not perform well on other styles of images (e.g., medical images). To prove this point, we show the results of supervised LDF \cite{ldf} and our unsupervised method on chest X-ray images \footnote{https://lhncbc.nlm.nih.gov/LHC-downloads/downloads.html} in Tab. \ref{tab:xray} and Fig. \ref{fig:visual}. It proves that SOD methods trained on existing SOD datasets perform poorly on X-ray images, while our unsupervised method can achieve significant performance without using extra human annotations for specific scenarios.
Third, many weakly-supervised methods leverage USOD methods as pre-processing or auxiliary loss, such as \cite{sal1, sal2}.
In addition, our method can be considered as a novel self-supervised learning paradigm.

\end{appendix}

\end{document}